\documentclass[letterpaper, 10 pt, conference]{ieeeconf}  

\IEEEoverridecommandlockouts                              

\usepackage{graphics} 
\usepackage{epsfig} 
\usepackage{mathptmx} 
\usepackage{times} 
\usepackage{amsmath} 
\usepackage{amssymb,mathtools}  
\usepackage{url}
\usepackage{booktabs} 
\usepackage{siunitx} 
\usepackage{float}
\usepackage{bm}
\usepackage{graphicx}
\usepackage{textcomp}
\usepackage{lipsum}

\title{\LARGE \bf
GMMCalib: Extrinsic Calibration of LiDAR Sensors using GMM-based Joint Registration
}

\author{Ilir Tahiraj$^{1, *}$, Felix Fent$^{1}$, Philipp Hafemann$^{1}$, Egon Ye$^{2}$, Markus Lienkamp$^{1}$
\thanks{$^{*}$ Corresponding author.}
\thanks{$^{1}$Authors are with the TUM School of Engineering and Design, Chair of Automotive Technology,
        Technical University of Munich. {\tt\small ilir.tahiraj@tum.de}.}%
\thanks{$^{2}$Egon Ye is with the TUM School of Computation, Information and Technology, Professorship of Cyber Physical Systems, Technical University of Munich.}
}

\usepackage{textcomp}
\usepackage{lipsum}
\usepackage{tikz}

\newcommand\copyrighttext{%
    \footnotesize \textcopyright{} This work has been submitted to the IEEE for possible publication. Copyright may be transferred without notice, after which this version may no longer be accessible.
}
\newcommand\copyrightnotice{%
    \begin{tikzpicture}[remember picture, overlay]
        \node[anchor=south, yshift=10pt] at (current page.south) {\fbox{\parbox{\dimexpr\textwidth-\fboxsep-\fboxrule\relax}{\copyrighttext}}};
    \end{tikzpicture}
}

\begin{document}

\maketitle
\copyrightnotice
\thispagestyle{empty}
\pagestyle{empty}

\begin{abstract}
State-of-the-art LiDAR calibration frameworks mainly use non-probabilistic registration methods such as Iterative Closest Point (ICP) and its variants. These methods suffer from biased results due to their pair-wise registration procedure as well as their sensitivity to initialization and parameterization. This often leads to misalignments in the calibration process. Probabilistic registration methods compensate for these drawbacks by specifically modeling the probabilistic nature of the observations. This paper presents GMMCalib, an automatic target-based extrinsic calibration approach for multi-LiDAR systems. Using an implementation of a Gaussian Mixture Model (GMM)-based registration method that allows joint registration of multiple point clouds, this data-driven approach is compared to ICP algorithms. We perform simulation experiments using the digital twin of the EDGAR research vehicle and validate the results in a real-world environment. We also address the local minima problem of local registration methods for extrinsic sensor calibration and use a distance-based metric to evaluate the calibration results. Our results show that an increase in robustness against sensor miscalibrations can be achieved by using GMM-based registration algorithms. The code is open source and available on GitHub\footnote[3]{https://github.com/TUMFTM/GMMCalib}.
\end{abstract}

\section{INTRODUCTION} \label{sec:introduction}
Autonomous robots fuse data from multiple sensors to understand the environment and detect objects. For robust sensor fusion, calibration is an essential task, i.e., sensors must be aware of their position and orientation relative to each other. The accuracy of this fusion process has a significant impact not only on object detection, but errors propagate throughout the functional chain, including planning algorithms and control functions involved in safety-critical tasks~\cite{fusion}. Calibration aims to ensure accurate alignment of sensor data. 
This work focuses on extrinsic calibration, which solves the spatial transformation between sensor frames.

For LiDAR sensor setups, this may require matching point cloud features in the respective sensor frames using point cloud registration. A common approach to point cloud registration is the ICP algorithm or its variants. These methods typically output a transformation matrix that describes the relative position and orientation between the point clouds, which can be used as the extrinsic calibration transformation. Although the ICP algorithm is efficient and simple, it has some drawbacks. It requires the spatial pose difference between the source and target point clouds to be small and is therefore sensitive to initialization and parameterization~\cite{Angel}. It also requires that one of the point clouds is defined as the reference frame, which can introduce a bias in the registration procedure. This becomes problematic when dealing with noisy real-world data and accordingly when performing extrinsic calibration. 

The goal of extrinsic calibration for many robotic applications is to estimate the transformation matrix as robustly as possible while maintaining local and global accuracy, i.e., the registration result performed on point clouds in the local neighborhood is also accurate at larger distances. In this context, robustness is referred to as maintaining accuracy without miscalibrations during the registration/calibration process. The risk of a registration outcome to return a local minimum solution is a concern for local registration methods such as ICP~\cite{deepgmr}.
\begin{figure}[!h]
      \centering
      \includegraphics[scale=0.25]{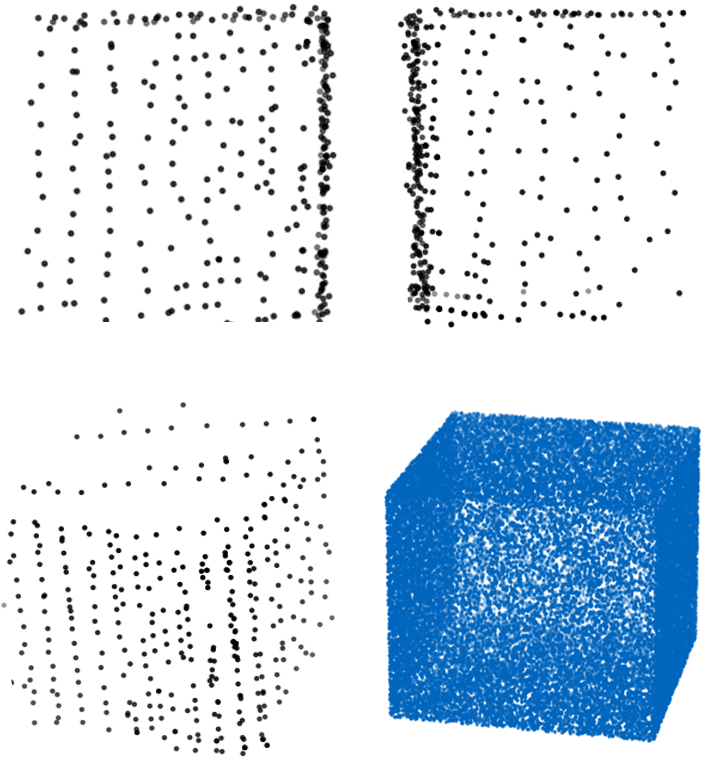}
      \caption{This figure illustrates the front (upper left), side (upper right) and pitched (lower left) view of the reconstructed calibration target colored in black and the ground truth point cloud (lower right) colored in blue.}
      \label{fig:geometric}
\end{figure}
However, among local registration methods, there are some algorithms that aim to find a more general solution than ICP by specifically considering noise and performing joint registration of many observations instead of pairwise registration~\cite{gmmreg, evangelidis}. In addition, these methods aim to avoid selecting one observation as the reference point cloud but instead estimate the latent geometric shape from which the observations have been drawn. These approaches formulate the registration problem probabilistically, often solving a Gaussian Mixture Model (GMM). The latent space solution of a GMM attempts to find the underlying distribution of the point cloud, thereby modeling both the characteristics of the data and the model from which the data was sampled. These techniques show great potential for the application in extrinsic sensor calibration. The main drawback is the computation time, which increases with the number of observations and the number of parameters of the GMM~\cite{hgmr}. The advantages are the probabilistic modeling of the noise during registration and the generation of a shape, which allows the use of geometric priors (Fig.~\ref{fig:geometric}). In this work, an extrinsic sensor calibration framework is introduced that implements a GMM-based joint registration algorithm to increase the robustness of the calibration process. We mitigate the computational drawbacks by using an arbitrary calibration target, thereby formulating the calibration problem as a reconstruction problem of the calibration target. Note, however, that this approach is not limited to the use of a calibration target and, given sufficient computing resources, can also be performed in a target-less fashion. To demonstrate the effectiveness of our approach  with both simulation and real-world experiments we compare the most common ICP algorithms used in the field of LiDAR calibration: Point-to-Point ICP \cite{besl}, Point-to-Plane ICP, Generalized-ICP (GICP) \cite{GICP}. The contributions of our work can be summarized as follows: 

\begin{itemize}
    \item To the best of our knowledge, we are the first to use a probabilistic GMM-based joint registration method for extrinsic sensor calibration
    \item Robust and accurate extrinsic sensor calibration only relying on the registration algorithm and no additional pre-processing or refinement effort and with significantly less miscalibrations
    \item With a geometric prior and the reconstruction of the calibration target, a plausibility check of the calibration result is possible
\end{itemize}

\section{RELATED WORK}
The field of extrinsic sensor calibration can generally be divided into two categories: Motion-based and Appearance-based sensor calibration. The former category aims to solve the spatial relationship between sensors by assuming that all frames follow a rigid body motion and solving the hand-eye problem \cite{horaud}. Appearance-based calibration solves the calibration problem by matching features in the environment, which can generally be described as a registration problem. In this section, we review works in the field of appearance-based calibration, with a focus on LiDAR sensor calibration, which use registration methods in various forms.

\subsection{Non-Probabilistic Registration Methods}\label{sec:nonprob}
The ICP algorithm is a non-probabilistic registration method and used in many appearance-based sensor calibration applications~\cite{lidarlink}. The most general variants used in extrinsic sensor calibration are the Point-to-Point, Point-to-Plane, and Generalized ICP. The Point-to-Point ICP algorithm uses direct point correspondences, which is robust when correspondences are known a priori. Zhang~\textit{et al}.~\cite{Zhang22} demonstrate this in a target-based approach, where point clouds are accumulated and the sphere center is estimated across multiple locations, followed by obtaining the transformation matrix using the Point-to-Point ICP algorithm. In a camera-to-LiDAR calibration framework presented by \cite{Yoon21}, a camera-based point cloud is reconstructed from images and aligned with a LiDAR-generated point cloud using the Point-to-Point ICP algorithm to initialize the relative pose between the sensors. In \cite{Angel}, a target-less 3-D LiDAR calibration framework is presented. They discuss the initialization and parameterization problems associated with using the ICP as a registration method, and increase robustness by introducing an iterative method for finding good initial point ICP initialization.

Point-to-Plane based algorithms are frequently employed in calibration tasks and have demonstrated good performance, particularly when applied to planar objects \cite{Li22, Kim21}. In \cite{Li22}, an algorithm for calibrating a depth camera to a 2D laser rangefinder using the Point-to-Plane ICP is introduced. They use an orthogonal trihedron as a target and perform calibration based on a single-shot observation. A more general optimization technique for plane fitting called IRLS is used in \cite{Kim21} to compensate for robustness and outlier issues in Point-to-Plane ICP. Jiao~\textit{et al}.~\cite{Jiao} present a target-less calibration framework for extrinsic calibration of a dual LiDAR system, utilizing planar surfaces in the environment. In a two-step optimization process, they  compute the alignment of detected planes and address local minima in the registration process. Their approach demonstrates an increase in robustness compared to relying solely on Point-to-Plane ICP. Hu~\textit{et al}.~\cite{Hu19} introduced a target-less extrinsic self-calibration method for LiDAR and stereo cameras. They reconstruct point clouds from stereo camera data and perform point cloud registration. The calibration approach uses an extended version of the classical Point-to-Point ICP algorithm, incorporating normal vectors in a triangle mesh representation rather than relying on point clouds. CROON~\cite{Wei22} is another target-less approach. The authors of \cite{Wei22} propose an efficient algorithm for LiDAR-to-LiDAR calibration utilizing an ICP variant named ICPN, inspired by the work of \cite{Serafin15}. 

In \cite{Heide18}, a target-less calibration framework for multiple 3D LiDAR sensors is introduced, utilizing the GICP algorithm~\cite{GICP}, where the sensor measurements from each LiDAR are subsequently merged into a calibrated point cloud through point cloud registration. LiDAR-Link~\cite{lidarlink} performs a two-step target-less calibration process for non-overlapping LiDARs, first a coarse initial alignment using a Motion-based approach and then a GICP registration on multiple point cloud pairs to jointly optimize for the extrinsic parameters.
\subsection{Probabilistic Registration Methods}\label{sec:prob}
There are many probabilistic registration methods that specifically address the handling of outliers and noise, as well as robustness in the registration process, which seem promising for extrinsic sensor calibration. EM-ICP~\cite{emicp} is a variant of the ICP algorithm that allows joint registration of multiple observations using the Expectation-Maximization (EM) algorithm \cite{emalgo}. The NDT-based algorithm presented in \cite{ndt} performs the joint registration on data in a voxel grid representation. While both the EM-ICP~\cite{emicp} and NDT~\cite{ndt} methods are fast and provide robust results, they do require the definition of a reference point cloud. GMM-based registration methods such as GMMReg~\cite{gmmreg} and JRMPC~\cite{evangelidis} also allow probabilistic registration of multiple point clouds. Advantageously, JRMPC does not require one of the observations to be defined as the reference point cloud, which is beneficial since its algorithm can return an optimized Gaussian Mixture Model representing an unbiased shape of the object. HGMR~\cite{hgmr} and MLMD~\cite{mlmd} focus on the efficiency of GMM-based registration methods, while \cite{deepgmr} provides a learning-based approach to GMM-based point cloud registration. 

In summary, the calibration frameworks presented in Section \ref{sec:nonprob} perform ICP registration or related methods to estimate the transformation between the sensors. However, these methods rarely perform direct registration, since ICP often lacks robustness or accuracy, thereby requiring additional algorithmic or pre-processing efforts to address the limitations of ICP algorithms. The probabilistic registration methods presented in Section \ref{sec:prob} aim to compensate for the drawbacks of the ICP algorithms, offering the potential to rely only on the registration for extrinsic sensor calibration.
Our work implements the more general JRMPC approach presented in \cite{evangelidis}, as it allows joint registration for sensor calibration and returns the reconstructed shape of the calibration target and thereby introducing a framework that incorporates a probabilistic registration method for extrinsic sensor calibration.

\section{METHODOLOGY}\label{sec:calibration}
To provide a good initial estimate for the following registration and calibration problem, the point clouds from the respective LiDAR sensors are aligned in a vehicle reference frame. As in most robotic platforms and especially in autonomous vehicles, the expected position and orientation of the sensors relative to the vehicle frame is given. This can be used to transform the point clouds to a common frame, resulting in an initial spatial alignment of the point clouds and, if existent, with a remaining misalignment representing the calibration error. For brevity, but without loss of generality, a dual LiDAR sensor setup will be considered for the remainder of the paper. This section describes the sensor calibration approach. Before describing the LiDAR-to-LiDAR calibration procedure, a brief introduction to the GMM-based joint registration method is provided. 

\subsection{GMM-based Joint Registration} \label{sec:gmm}
Let $N = N_{L_1} + N_{L_2}$ (with $N_{L_1}=N_{L_2}$) be the number of concatenated point sets observed from LiDAR frames $\{L_{1}, L_{2}\}$. $\mathbf{O}_{i} = [\mathbf{x}_{i1} \dots \mathbf{x}_{ik} \dots \mathbf{x}_{iN_{i}}] \in \mathbb{R}^{3 \times N_i}$ are $N_i$ points belonging to the point set $i$. 
Despite the fact that the points come from different sensors, the assumption remains that the observed points in $\mathbf{O}_{i}$ are generated from the same mixture model. Therefore, we will perform the joint registration on the union of points $\mathbf{O} = \{\mathbf{O}_i\}^{N}_{i = 1}$.  In our case, however, it is important to assign the optimized parameters to a specific sensor frame. This knowledge is necessary to identify which transformations are applied to each point set $\mathbf{O}_{i}$ during the registration process and to compute the spatial relationship between the corresponding sensors. 

The goal of the joint registration is to find $N$ transformations in the observer frames to the calibrated frame. The parameters of the GMM-based registration method are the model parameters $\bm{\Theta}_1 = \left \{p_m,  \bm{\mu}_m, \bm{\Sigma}_m \right \}_{m=1}^{M}$ and the set of transformations $\bm{\Theta}_2 = \left \{ \mathbf{R}_i, \mathbf{t}_i \right \}_{i=1}^{N}$. The first parameter set describes the Gaussian Mixture Model, which, in the converged state, is a point cloud of size $M$ with a covariance $\bm{\Sigma}_m  \in \mathbb{R}^{3 \times 3}$ for each mixture component $\bm{\mu}_m  \in \mathbb{R}^{3}$ and a designated probability $p_m$ of a point being sampled from that underlying Gaussian distribution. The second set provides the transformations consisting of $\mathbf{R}_i \in \mathbb{R}^{3 \times 3}$ and $\mathbf{t}_i \in \mathbb{R}^{3}$ to that underlying model. The parameters $\bm{\Theta} = \left \{ \bm{\Theta}_1, \bm{\Theta}_2 \right \}$ are jointly optimized using the EM algorithm. The correspondences between the Gaussian Mixture components $\bm{\mu}_m$ and the observed points $\mathbf{x}_{ik}$ in $\mathbf{O}_i$ are referred to as the latent variables $Z_{ik}$ and it can be understood as $Z_{ik} = m$ if $\mathbf{x}_{ik}$ is a correspondence point to component $m$. The expected log-likelihood function is maximized with respect to the latent variable as described in \cite{evangelidis}: 

\begin{equation}
    f(\bm{\Theta}|\mathbf{O}, Z) = \mathbb{E}_Z \left[log P(\mathbf{O}, Z | \mathbf{O}, \bm{\Theta} ) \right].
\end{equation}

\subsection{LiDAR-to-LiDAR Calibration} \label{sec:lidarcalibration}
\begin{figure}[thpb]
      \centering
      \includegraphics[scale=0.21]{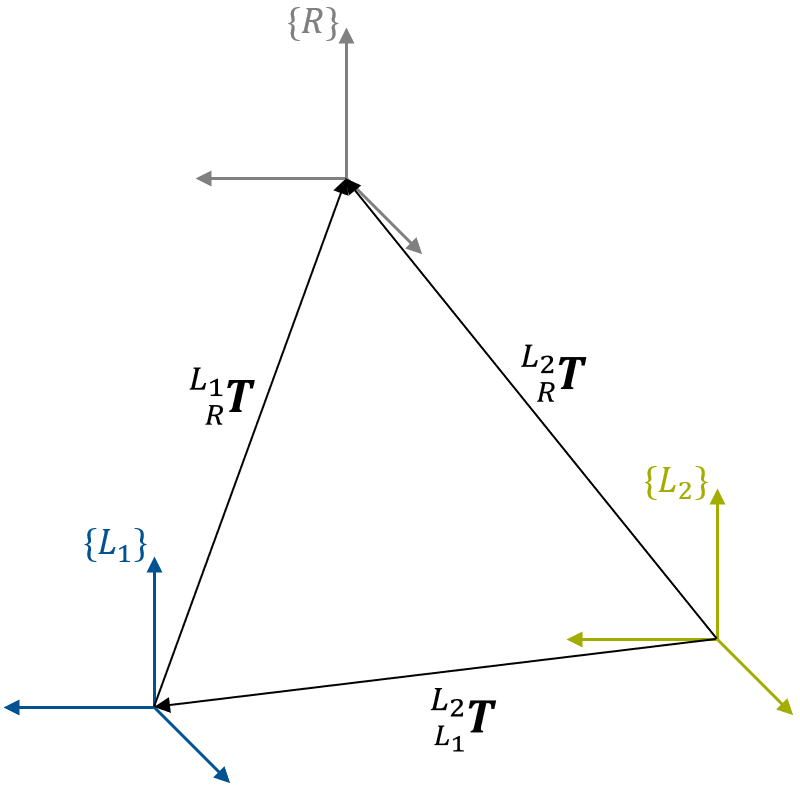}
      \caption{Spatial relationship between the LiDAR sensor frames and the calibrated frame with $\{L_1\}$ and $\{L_2\}$ representing the sensor coordinate frames and $\{R\}$ representing the calibration frame.}
      \label{fig:frames}
   \end{figure}

The goal of extrinsic calibration is to find the rotation matrix and the translation vector between different sensor coordinate frames. The rotation and translation are expressed in
homogeneous transform notation according to \cite{craig}. As shown in Fig. \ref{fig:frames}, $\{L_1\}$ and $\{L_2\}$ represent the sensor coordinate frames of the two LiDAR sensors, while $\{R\}$ describes the coordinate frame of the aligned point clouds. Since the registration is performed with the point clouds expressed in the vehicle frame, Fig. \ref{fig:frames} illustrates the relationships after the transformation into the vehicle frame as described in \ref{sec:gmm}. $\{R\}$ is introduced, because due to the nature of GMM-based registration, both sensor frames are transformed into an arbitrary reference frame. This differs from ICP registration, which directly solves for the transformation from the defined source point cloud to the target point cloud. In such a case, $\{L_2\}$ is defined as the source frame and $\{L_1\}$ as the reference frame.

Let $\prescript{L_2}{L_1}{\mathbf{T}}$ be the true transformation matrix describing the calibration error between $\{L_1\}$ and $\{L_2\}$. Given $\prescript{L_1}{R}{\mathbf{T}}$
and $\prescript{L_2}{R}{\mathbf{T}}$, which are the transformations of the sensor frames to the common reference frame $\{R\}$, respectively, the extrinsic calibration transform can be recovered as follows: 

\begin{equation} \label{eq:true}
    \prescript{L_2}{L_1}{\mathbf{T}}  = \prescript{L_2}{R}{\mathbf{T}}\cdot \prescript{L_1}{R}{\mathbf{T}}^{-1}.
\end{equation}

Since we will compare different algorithms, we use the more general notation $\prescript{L_2}{L_1}{\mathbf{T}_{r,j}}$ to describe the resulting calibration matrix. As mentioned in the previous subsection, we have to keep track of the parameters corresponding to the respective sensor frames. The resulting calibration matrix is then computed as follows:  

\begin{equation} \label{eq:gmm}
    \prescript{L_2}{L_1}{\mathbf{T}_{r,j}}  = \prescript{L_2}{R}{\mathbf{T}_{r,j}}\cdot \prescript{L_1}{R}{\mathbf{T}_{r,j}}^{-1},
\end{equation}

where $\prescript{L_1}{R}{\mathbf{T}_{r,j}} = \{\mathbf{T}_i\}^{N_{L_1}}_{i = 1}$ is the set of transformations from frame $\{L_1\}$ and $\prescript{L_2}{R}{\mathbf{T}_{r,j}} = \{\mathbf{T}_i\}^{N}_{i = N_{L_2}}$ from frame $\{L_2\}$. The subscript $j = 1, \dots, N/2$ indexes the $N/2$ calibration transformations returned from the joint registration of the corresponding observations and $r$ describes the corresponding algorithm. 

\section{EXPERIMENTAL DESIGN}
In this section, we present the experimental setup to investigate the calibration of the LiDAR setup using cubic calibration targets. Our study combines both simulation-based and real-world analysis to evaluate the robustness and accuracy of the calibration approach. The point cloud measurements were performed while the vehicle was stationary. In this way, no motion and no time synchronization between the sensors is required, which reduces additional sources of errors. The sensors placements were on front left and front right position relative to the vehicle origin both in simulation as well as real-world environment.

\subsection{Simulation}
\begin{figure}[thpb]
  \centering
  \includegraphics[scale=0.20]{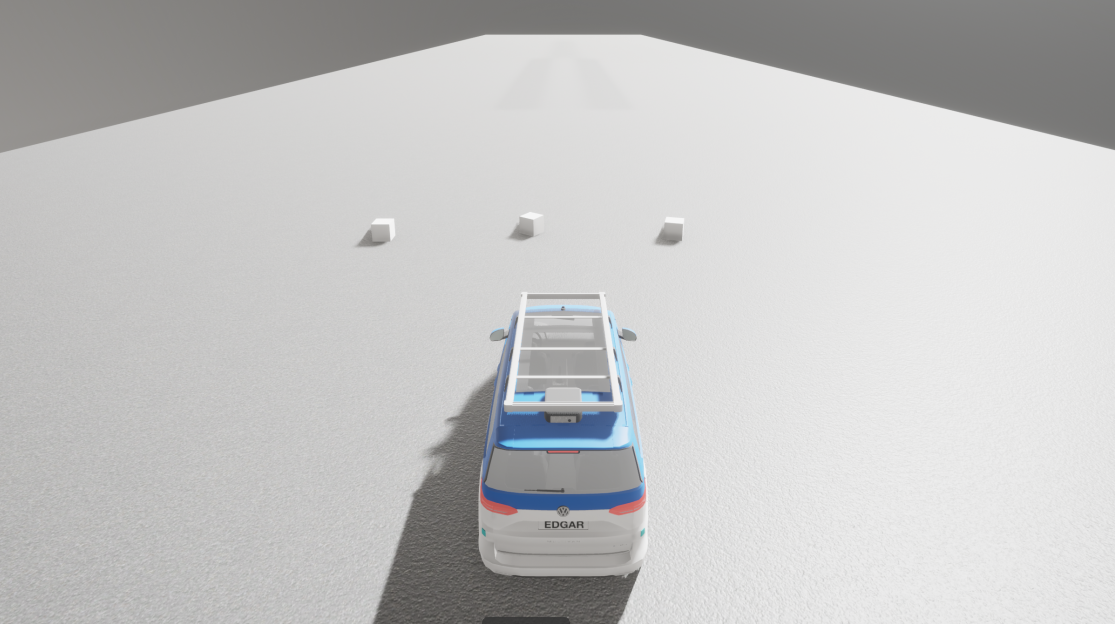}
  \caption{A CARLA simulation environment showing three cubic calibration targets with edges of 0.5 meters in close proximity. The digital twin of the EDGAR~\cite{edgar} research vehicle is placed at the origin of the world.}
  \label{fig:carla}
\end{figure}
The simulation environment was created using the CARLA simulator \cite{carla} and was designed to focus only on the aspects relevant to our calibration approach: The road was modeled as a flat surface that can be detected by the sensor models to also generate ground level points. Apart from adjusting the lighting for visual purposes, no additional weather conditions were simulated (Fig. \ref{fig:carla}). The sensor positions were chosen according to the digital twin of the research vehicle EDGAR. Table \ref{tab:comparison} shows the technical attributes of the sensor model used for both rotating LiDARs in the simulation experiments. Three cubes, oriented differently in the environment, were placed in front of the vehicle in the overlapping field of view (FoV) of both sensors.

\begin{table}[h]
\caption{Sensor characteristics}
\label{tab:comparison}
\centering
\begin{tabular}{lcc}
\toprule
\textbf{Attribute} & \textbf{Simulation} & \textbf{Real-World} \\
\midrule
Horizontal Field Of View [°] & 360 & 360 \\
Vertical Field Of View [°] & 25 & 45 \\
Range [m] & 50 & 120 \\
Channels [-] & 50 & 128 \\
Sensor Rate [Hz] & 10 & 10  \\
Precision (1-20 m) [m] & $\pm$ 0.01 & $\pm$ 0.01 \\
\bottomrule
\end{tabular}
\end{table}

\subsection{Real World}
\begin{figure}[thpb]
  \centering
  \includegraphics[scale=0.25]{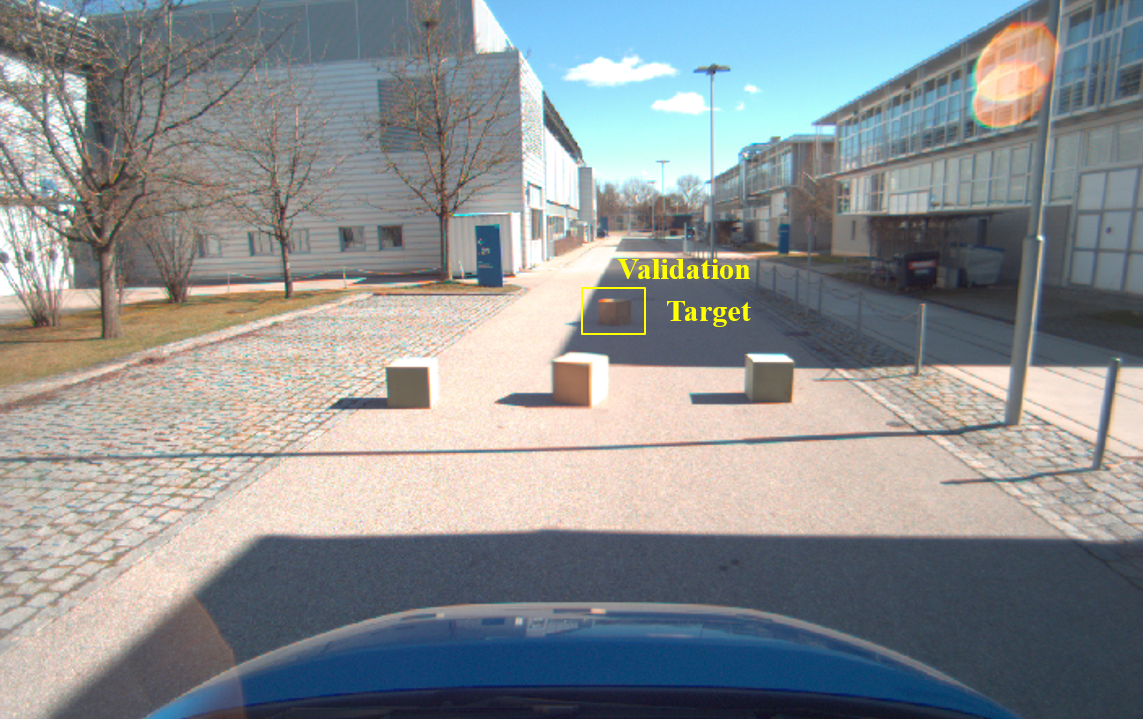}
  \caption{Real-world experimental setup with three cubic calibration targets with edges of 0.5 m placed on a regular road. Another target is placed at a distance of about 16 m.}
  \label{fig:edgar_real}
\end{figure}
The real-world experiments were performed to validate the results obtained from the simulation. The tests were performed similar to the simulation setup with three cubes positioned in front of the vehicle at 10 m distance and in the FoV of both sensors. An additional cube is placed at a distance of about 16 m for validation purposes (see Fig. \ref{fig:edgar_real}). Two Ouster OS1-128 LiDARs were mounted on EDGAR in a way to ensure an overlapping FoV. The technical details of the sensors are also listed in Table \ref{tab:comparison}. No significant weather conditions were observed during the experiments.

\section{RESULTS}
This section presents the simulation and real-world results of the proposed calibration approach. A sequence of measured point clouds from the conducted experiments is used to evaluate the performance and robustness of the LiDAR calibration framework. First, the evaluation of the algorithms is described before looking at the simulation and real world results. 
\subsection{Evaluation}
The simulated environment allows us to perform a comprehensive evaluation of the calibration results by means of a ground truth calibration error as depicted in Fig. \ref{fig:frames}. We simulated 100 calibration errors, where the point sets $\{\mathbf{O}_i\}^{N}_{i = N_2}$ are randomly transformed by a roll, pitch, and yaw angle error between $\pm 3°$ as well as a translation error of $\pm 0.1~m$, respectively. As mentioned in Section~\ref{sec:lidarcalibration}, $\prescript{L_2}{L_1}{\mathbf{T}}$ represents the true calibration error between the LiDAR sensor frames $\{L_1\}$ and $\{L_2\}$, which is known in the simulation environment. The transformation result of a registration algorithm provides a transformation of the frame $\{L_2\}$ that coincides with $\{L_1\}$ only if the registration algorithm finds the global minimum, namely the transformation error $\prescript{L_2}{L_1}{\delta \mathbf{T}} = \mathbf{I}$ being the identity matrix or in other words, zero roll $\delta \phi$, pitch $\delta \theta$, and yaw $\delta \psi$ angle errors, and zero translation component errors $\delta x$, $\delta y$, and $\delta z$. 
\begin{figure}[thpb]
  \centering
  \includegraphics[scale=0.16]{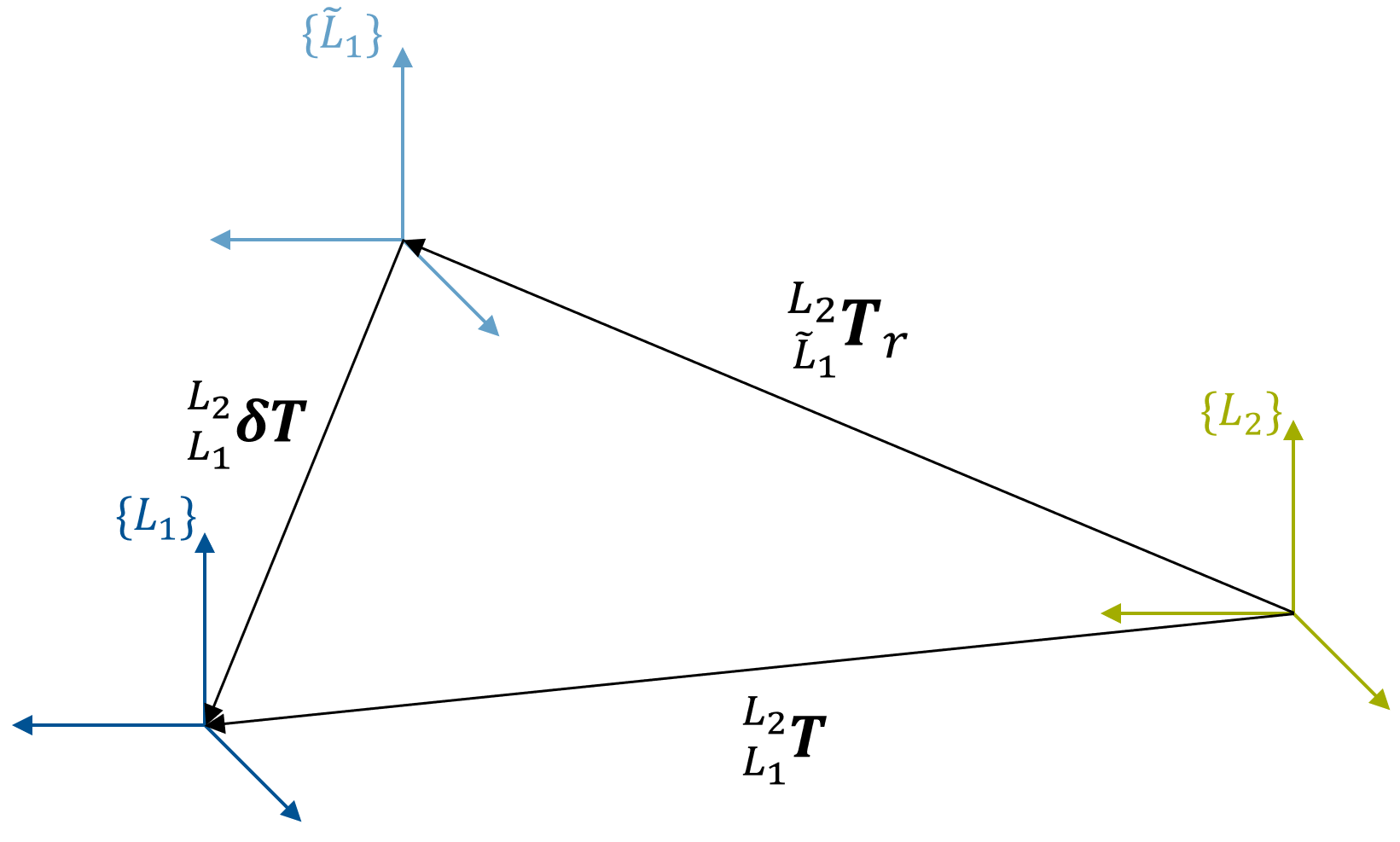}
  \caption{Spatial relationship between the LiDAR sensor frames and $\{\widetilde{L}_1\}$ with $\{L1\}$ and $\{L2\}$ representing the sensor frames and $\{\widetilde{L}_1\}$ represents the erroneously transformed frame.}
  \label{fig:frameserror}
\end{figure}

This relationship is valid under the condition that gimbal lock does not occur. For this reason, we continue with the assumption that the calibration errors as well as the rotations involved during the registration process are less than 90 degrees, which can be assumed for the former errors and the registration process can be tracked to satisfy the latter constraint. Since neither the GMM-based registration nor the ICP is guaranteed to converge to a global minimum, the resulting transformation may transform the point clouds into a $\widetilde{{L_1}}$ frame, as shown in Fig. \ref{fig:frameserror}.
Under the previous assumptions, we can compute the transformation error of the registration algorithms as follows: 
\begin{equation}\label{eq:transformation}
     \prescript{L_2}{L_1}{\delta \mathbf{T}_{j}} =  \prescript{L_2}{\widetilde{L}_1}{\mathbf{T}_{r,j}^{-1}} \cdot \prescript{L_2}{L_1}{\mathbf{T}}
\end{equation}
The index $r$ represents the registration algorithm and $j$ refers to the corresponding observation pair from which the calibration transform is computed. Note that the joint registration returns a calibration matrix for each observed point cloud pair. To compare the GMM-based transformations with the ICP algorithms, we run each ICP registration algorithm on the same point cloud pairs. This results in four algorithm-related sets of $N/2$ transformations. 

However, looking only at $\prescript{L_2}{L_1}{\mathbf{\delta T}}$ can lead to an ambiguous interpretation of accuracy or robustness. While the errors between the transformations of each algorithm may differ by direct numerical comparison, the final registration result may still be adequate. Given the characteristics of registration algorithms, they provide 6 degrees of freedom for point cloud alignment, so this metric alone is not sufficient to evaluate the calibration performance. Therefore, we also introduce a distance-based metric. Given the ground truth solution $\prescript{L_2}{L_1}{\mathbf{T}}$, we can compute the mean distance error per point with known point correspondences. The Equation (\ref{eq:distance}) returns the mean distance error in $x$, $y$, and $z$ between the points in the observation $\mathbf{O}_j = \{\mathbf{O}_i\}^{N}_{i = N_2}$ after being transformed by $\prescript{L_2}{L_1}{\mathbf{T}}$ and $\prescript{L_2}{\widetilde{L}_1}{\mathbf{T}_{r,j}^{-1}}$, respectively:
\begin{equation}\label{eq:distance}
    \mathbf{\delta X}_r = \left[ \frac{1}{N_j} \sum_{k=1}^{N_j} \left(\prescript{L_2}{L_1}{\mathbf{T}} \cdot \mathbf{x}_{jk} - \prescript{L_2}{\widetilde{L}_1}{\mathbf{T}_{r,j}^{-1}} \cdot \mathbf{x}_{jk} \right) \right]_{j=1}^{N_2}
\end{equation}

with $\mathbf{\delta X}_r \in \mathbb{R}^{3 \times N_{L_2}}$. This distance metric is computed on the local point clouds used for the registration algorithms. 
In the final step, the distance error is computed over the entire point cloud of the scene to evaluate the calibration transformation on a more global scale, i.e. points at larger distances that were not used in the optimization procedure.

Similar to the distance metric in Equation (\ref{eq:distance}), we can make use of the known point correspondences after transformations. The second distance metric uses the L2 norm of the point errors between the transformed point clouds. We introduce a second distance metric to evaluate the calibration results at larger distances, to indicate whether the algorithms find solutions that are robust and accurate on a more global scale. 
\subsection{Simulation}
\begin{figure}[h]
  \centering
  \includegraphics[scale=0.21]{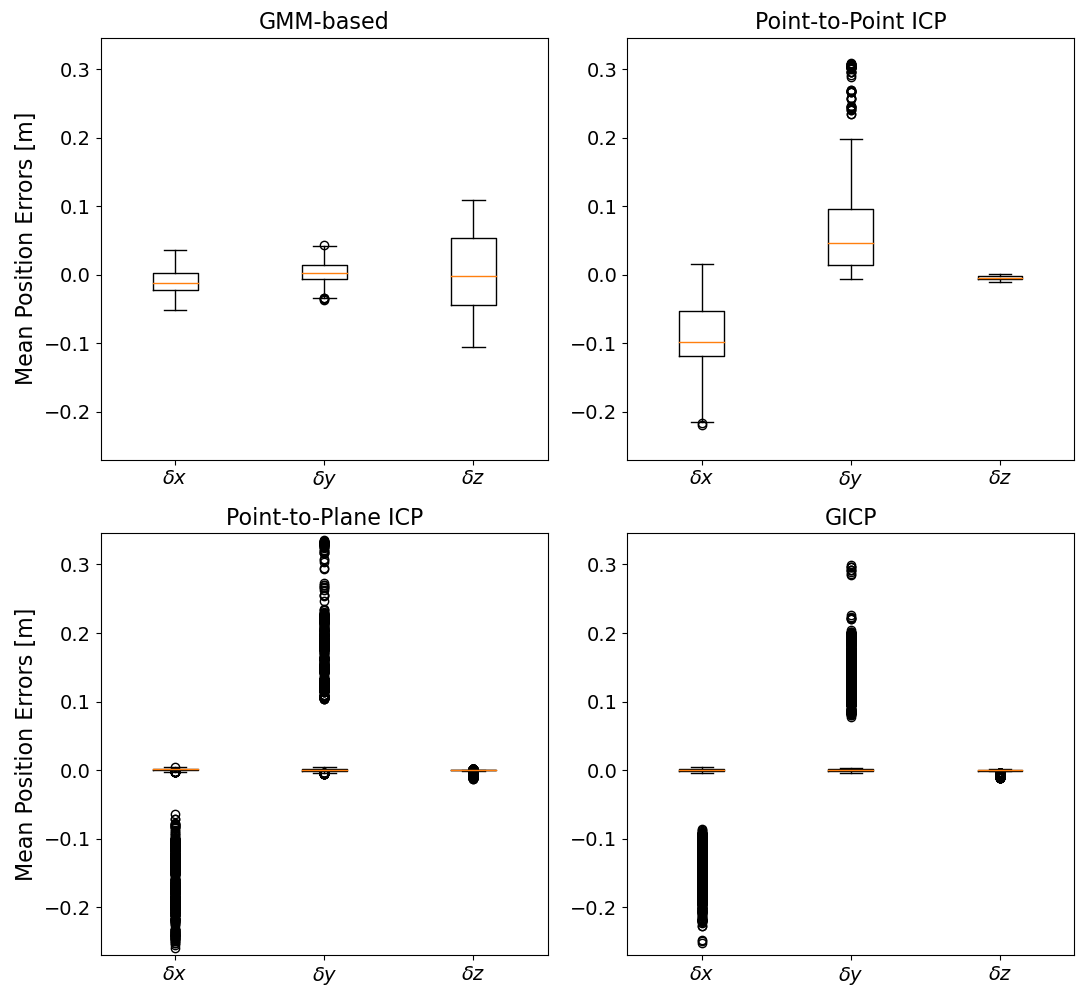}
  \caption{Evaluation of the Mean Position Errors $\mathbf{\delta X}_r$ computed according to Equation (\ref{eq:distance}). The plots describes the average distance error between the point clouds transformed by $\prescript{L_2}{L_1}{\mathbf{T}}$ and $\prescript{L_2}{\widetilde{L}_1}{\mathbf{T}_{r,j}^{-1}}$ in $x$, $y$, and $z$ directions.}
  \vspace{-5pt}
  \label{fig:res_robustness_sim}
\end{figure}

\begin{figure}[!h]
  \centering
  \includegraphics[scale=0.19]{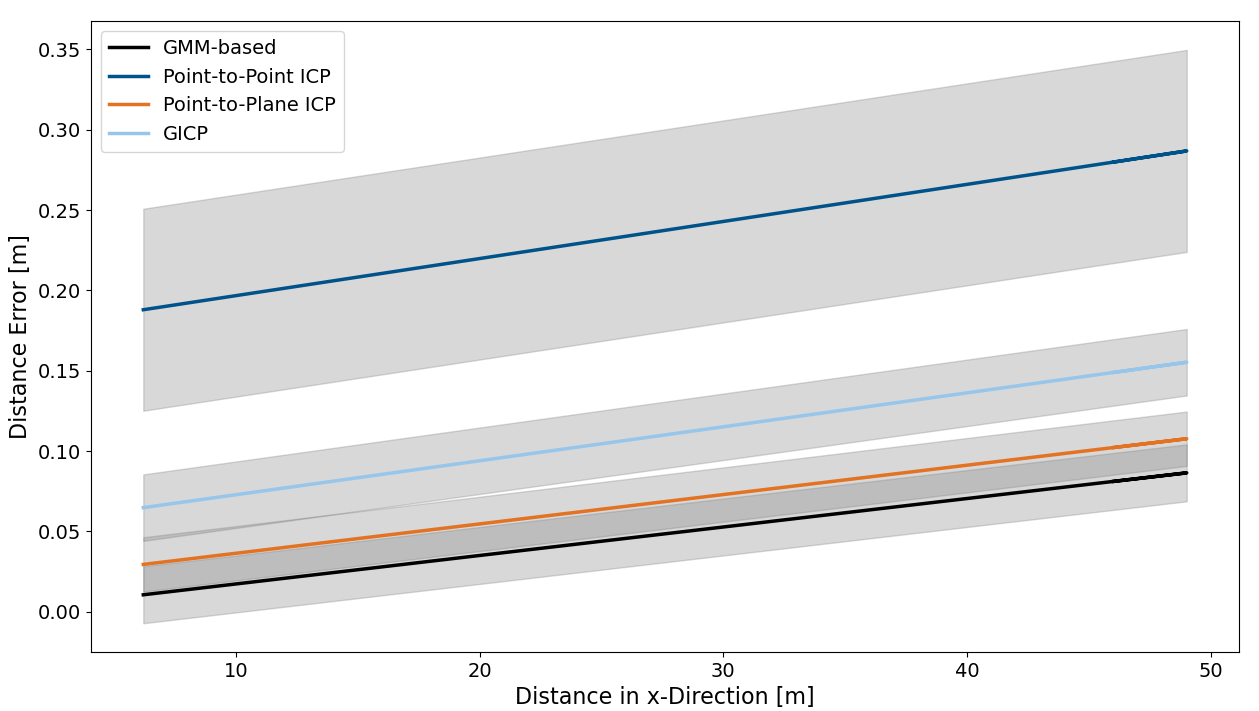}
  \caption{The distance errors (L2 norm) between the point clouds transformed by $\prescript{L_2}{L_1}{\mathbf{T}}$ and $\prescript{L_2}{\widetilde{L}_1}{\mathbf{T}_{r,j}^{-1}}$ over the distance in x-direction are computed. The lower and upper bounds illustrate the standard deviations of the distance error.}
  \label{fig:res_global_sim}
  \vspace{-3pt}
\end{figure}

The results shown in this section were obtained with $N_{L_1} = N_{L_2} = 106$ resulting to a total number of $N = 212$ observations and $\mathbf{O}_i$ with an average point cloud size of $\approx 1850$ points. The total point cloud size for each observation consisted of $\approx 12500$ points. The $M = 400$ mixture components were initialized as cubes and placed at random positions in the periphery of the observations. The calibration results for this data set are presented in Table~\ref{tab:res_transform_sim}, Fig.~\ref{fig:res_robustness_sim} and Fig.~\ref{fig:res_global_sim}. 
\begin{table}[!h]
    \centering
    \caption{Mean Euler Angle [in rad] and Translation [in m] Errors}
    \label{tab:res_transform_sim}
    \begin{tabular}{ccccccc}
        \toprule
         Algorithm & $\delta \phi$ & $\delta \theta$ & $\delta \psi$ & $\delta x$ & $\delta y$ & $\delta z$\\
        \midrule
        GMM       & 0.0033  & 0.0036  & \textbf{0.0020}            & \textbf{0.015} & \textbf{0.027}   & 0.018\\
        Point ICP & 0.0013  & 0.0011  & 0.0269            & 0.088          & 0.297            & 0.013\\
        Plane ICP & \textbf{0.0004}  & \textbf{0.0005}  & 0.0078            & 0.034          & 0.075            & \textbf{0.006}\\
        GICP      & 0.0006  & 0.0007  & 0.0139            & 0.072          & 0.179            & 0.009\\
        \bottomrule
    \end{tabular}
\end{table}
Table~\ref{tab:res_transform_sim} shows the mean Euler angle and translation error using Equation (\ref{eq:transformation}) over all $\prescript{L_2}{L_1}{\delta \mathbf{T}_{j}}$. The transformations are split into angular and translational parts, and the smallest errors are highlighted in bold. The results explain the ambiguity mentioned in the previous subsection. In the case of Point-to-Point ICP with a larger yaw error, a correlation with the translation error in y direction is observed. In other words, the translation error of $\delta y = 0.297~m$ compensates for the yaw error, and therefore still resulting in visually sufficient registrations. Similarly, the $\delta z$ error for the GMM-based calibration result provides a compensation of bigger errors in roll and pitch angles. Overall, the GMM-based and Point-to-Plane calibrations have accurate transformations, with the GMM-based approach being more accurate in translation and the Point-to-Plane ICP more accurate in the angular components.
In summary, for all registration methods, the correlation between translation and angular errors indicates a compensating effect rather than a calibration error. The transformation error metric cannot describe the underlying calibration uncertainties, so the calibration performance is further investigated using the distance metric shown in Fig. \ref{fig:res_robustness_sim}. The box plots are generated by evaluating Equation (\ref{eq:distance}) for the 100 randomly sampled calibration errors. For each calibration error $N_2$ registration results are plotted leading to a total number of $\approx 10000$ data points. The GMM-based approach, as well as Point-to-Plane ICP and GICP, indicate a high accuracy regarding the extrinsic calibration, with the two ICP algorithms being slightly more accurate in z-displacement than the GMM-based algorithm. The Point-to-Point ICP does not provide accurate or robust calibration results. The main aspect that Fig. \ref{fig:res_robustness_sim} highlights is the robustness of the algorithms. For each of the ICP variants, there are cases of miscalibration, shown as outliers, while the GMM-based approach has a slightly larger standard deviation compared to Point-to-Plane and GICP but significantly fewer miscalibrations. Finally, the distance-based metric is evaluated on the entire point cloud to determine the accuracy of the algorithms on the larger periphery, indicating which calibration approach has better generalization characteristics. To describe this accuracy in a simple relation, the L2 norm error between the point clouds transformed by $\prescript{L_2}{L_1}{\mathbf{T}}$ and the mean transformation of $\prescript{L_2}{\widetilde{L}_1}{\mathbf{T}_{r,j}^{-1}}$ is computed. Plotting all distance errors would result in a scattered point cloud. 
For illustration and comparability, a linear relationship is fitted through the points representing the distance error of each approach. The resulting function as well as the lower and upper bounds of the distance errors over the distance in the x-direction are shown in Fig. \ref{fig:res_global_sim}. The GMM-based approach outperforms the ICP approaches. The monotonically increasing relationship of the distance error indicates an angular calibration error. The function offsets curve indicate a translation error.
\subsection{Real World}
\begin{figure}[h]
  \centering
  \includegraphics[scale=0.25]{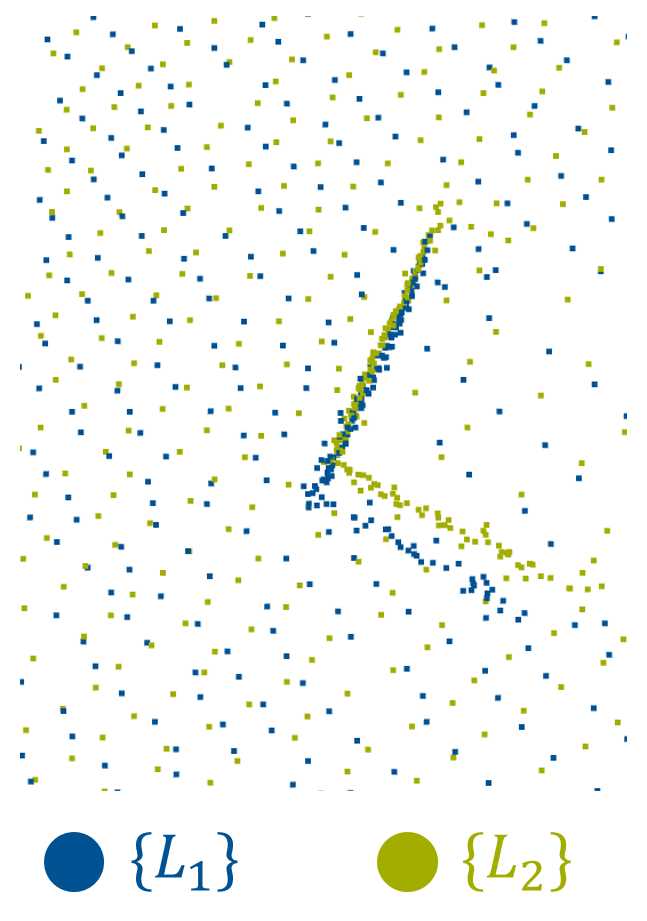}
  \caption{The initial calibration error between the two LiDAR sensors in the real-world experiments. The center Cube is shown in this Figure.}
  \vspace{-10pt}
  \label{fig:calerror}
\end{figure} 
The number of point clouds in the real world experiments were $N_{L_1} = N_{L_2} = 104$ leading to a total number of $N = 208$ observations with $\mathbf{O}_i$ on average consisting of $\approx 1400$ points. 
\begin{figure*}[!h]
  \centering
  \includegraphics[width=\textwidth, scale=0.8]{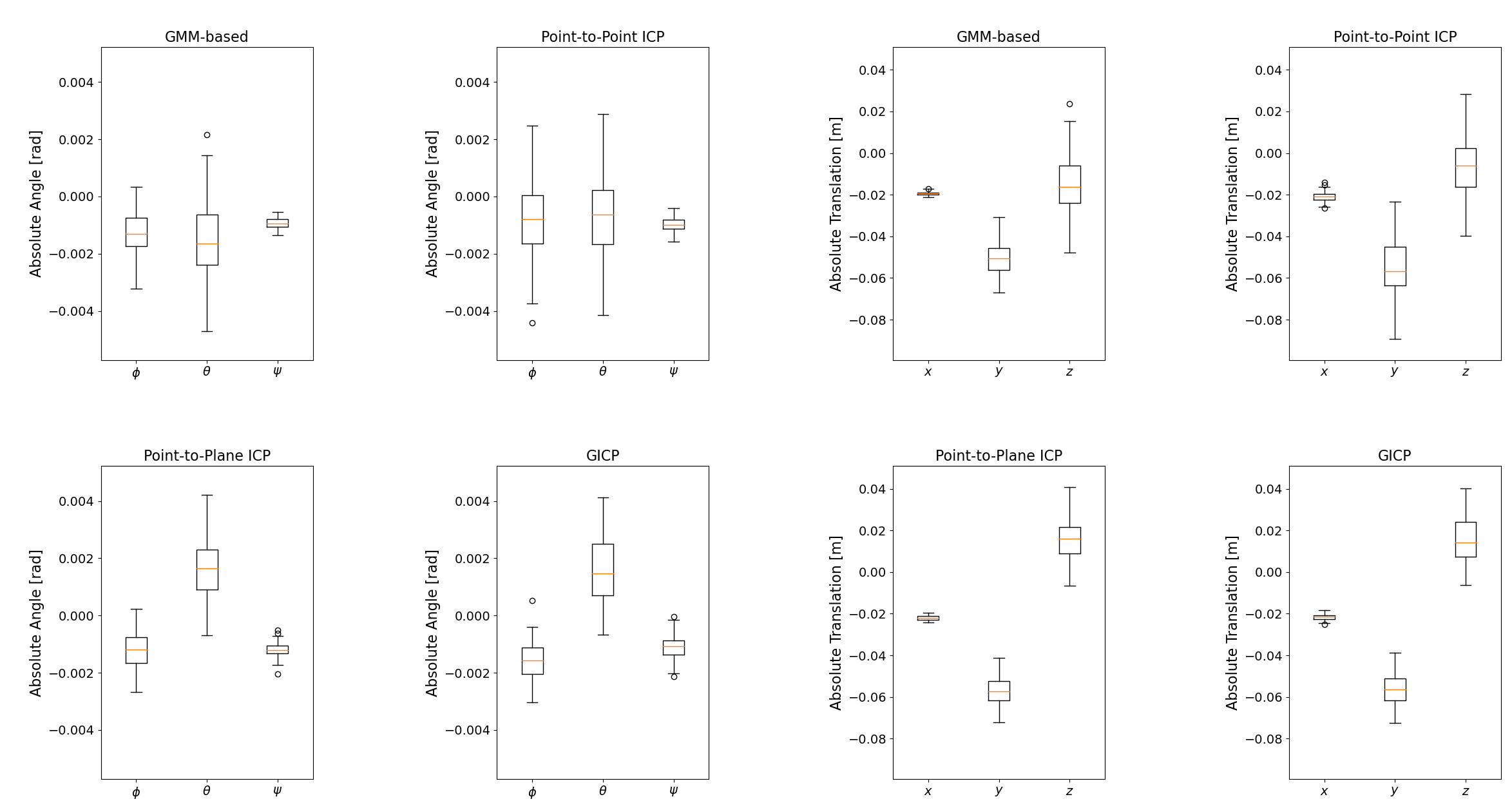}
  \caption{This plot provides the Euler angles and translation components of the calibration matrices obtained from the registration algorithms. The four left plots illustrate the Euler angles of the different algorithms. The four right plots are assigned to the absolute translation components.}
  \vspace{-10pt}
  \label{fig:res_robustness_real}
\end{figure*}
The mixture components were initialized according to the simulated experiments. The entire point clouds consisted of $\approx 131000$ points. In contrast to the simulated experiments, no ground truth of the transformation is provided and also no specific calibration procedure was performed prior to the real experiments, resulting in a relatively high calibration error between the two LiDARs by default. The bird's eye view of the central cube is shown in Fig.~\ref{fig:calerror}.
To build on the previous robustness evaluation, the box plots in Fig.~\ref{fig:res_robustness_real} show the actual transformation values. In terms of the transformation errors, a similar picture emerges with respect to the z-displacement. The GMM-based approach shows a higher standard deviation for the z-translation, while the Point-to-Plane and GICP algorithms have a close to equal z-component value. 
Note that, while the Point-to-Plane and GICP have an opposing z-direction, the pitch angle $\theta$ also opposes the pitch angle of the GMM-based approach, indicating a compensatory effect. 
\begin{figure}[!h]
  \centering
  \includegraphics[scale=0.33]{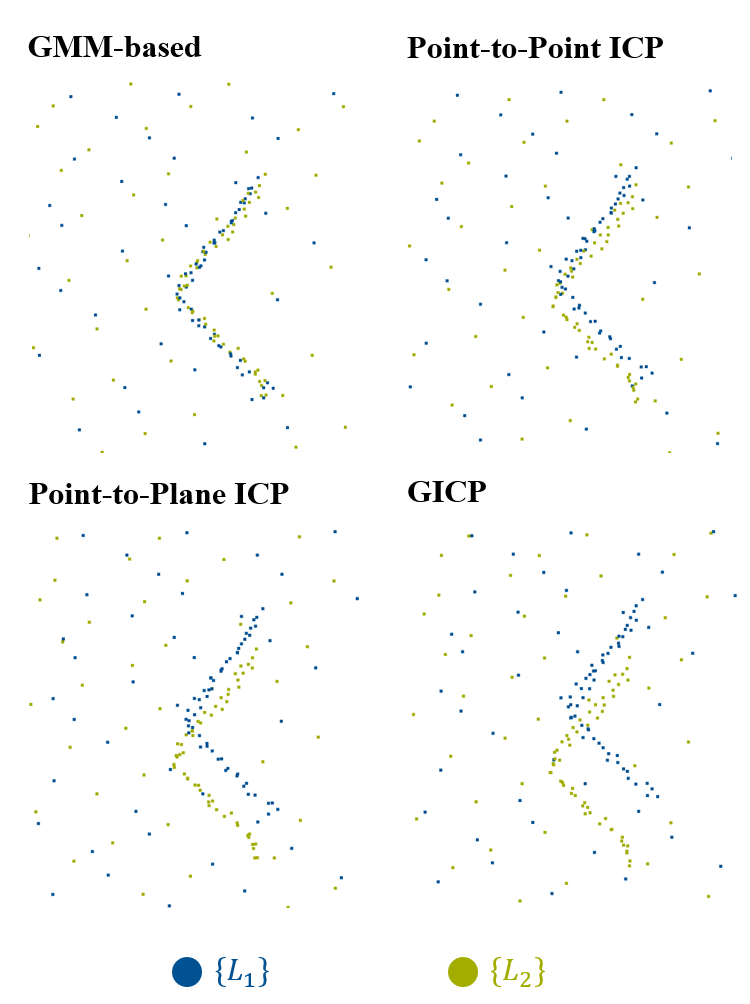}
  \caption{Visual comparison the the calibration results in real-world experiments for the cubic validation target. The GMM-based calibration procedure outperforms the ICP algorithms.}
  \vspace{-12pt}
  \label{fig:res_visual}
\end{figure}
However, following the same procedure as in the simulation environment and choosing the mean transformation of each algorithm to evaluate the algorithm on the cube placed at a larger distance, we see larger visual deviations between the GMM-based and the ICP algorithms in Fig. \ref{fig:res_visual}. Overall, given the qualitative calibration error in Fig. \ref{fig:calerror}, the different approaches share that the calibration error is compensated mostly via translation in x and y direction (shown in the right four plots). While from Fig. \ref{fig:res_robustness_real}, the differences between the algorithms seem small, Fig. \ref{fig:res_visual} indicates that the registration of the GMM-based approach in these specific real world experiments is more accurate and robust. 
\section{DISCUSSION}
The key aspect of this work is to highlight the robustness of choosing a data-driven approach to extrinsic sensor calibration. In order to use registration algorithms as standalone calibration approaches without further refinement or optimization steps, some important considerations are necessary. First, due to the fact that the registration algorithms evaluated in this work only solve for a local solution, the output in the form of a transformation matrix does not necessarily provide an intuitive answer. Relatively large translations in the calibration matrix may be due to high angular errors and can be considered compensatory parameters rather than descriptions of the true relative position and orientation between sensors. An additional and more comprehensive distance metric was introduced to evaluate robustness. Overall, the real-world results support the findings from the simulation experiments and demonstrate the potential of the GMM-based approach for extrinsic sensor calibration. Furthermore, the approach shows accuracy and significantly higher robustness compared to the ICP approaches in calibrating LiDAR sensor arrays. We observed that the ICP algorithm is more accurate in terms of z-displacement, as the ground level points were left in the observations. In the tests conducted during the development phase, removing them resulted in a significant decrease in the robustness of the ICP algorithms. Since the ground level points provide critical understanding in the calibration process, we decided to keep them in the experiments. Based on the results, the ICP algorithm seems to value the points more highly than the GMM-based algorithm. A limiting factor of using the joint registration approach for sensor calibration is the computation time. We reduced this effect by choosing specific targets, but for our experiments the computation time was $\approx$ 10 minutes on one computation unit. This can be reduced by parallelization, however, GMMCalib becomes particularly valuable with more data, so this effect can only be mitigated to a certain extent. In general, as an automatic offline approach, computational time is not necessarily the highest priority. Another advantage of our approach is the possibility to use a geometric prior as a plausibility check. The joint registration algorithm returns a reconstructed shape that can be compared to a point cloud generated from the CAD model.

\section{CONCLUSION \& FUTURE WORK}
This work presented GMMCalib, an extrinsic sensor calibration approach that uses a joint registration algorithm to obtain the spatial relationship of LiDAR sensors. It was shown that GMMCalib is a robust and accurate offline calibration approach, and with the proven reliability, it can be automatically performed on a robotic or autonomous driving platform. We proved that our approach overcomes some of the limitations of ICP algorithms in sensor calibration and demonstrated the potential of a more data-driven approach. Two main branches in the field of sensor calibration can be further investigated. First, the use of GMMCalib could be promising for a sensor system with a non-overlapping FoV, as the joint registration algorithms presented in this paper are dedicated to handle different perspectives of objects more accurately and robustly. Second, geometric priors can be used not only as a plausibility check but also as a constraint to optimize both the extrinsic and, especially in real-world scenarios, the intrinsic properties of LiDAR sensors. The geometric prior and the resulting reconstructed shape can implicitly serve as a ground truth for calibration.  


\bibliography{GMMCalib}

\begin{thebibliography}{10}
\providecommand{\url}[1]{#1}
\csname url@rmstyle\endcsname
\providecommand{\newblock}{\relax}
\providecommand{\bibinfo}[2]{#2}
\providecommand\BIBentrySTDinterwordspacing{\spaceskip=0pt\relax}
\providecommand\BIBentryALTinterwordstretchfactor{4}
\providecommand\BIBentryALTinterwordspacing{\spaceskip=\fontdimen2\font plus
\BIBentryALTinterwordstretchfactor\fontdimen3\font minus
  \fontdimen4\font\relax}
\providecommand\BIBforeignlanguage[2]{{%
\expandafter\ifx\csname l@#1\endcsname\relax
\typeout{** WARNING: IEEEtran.bst: No hyphenation pattern has been}%
\typeout{** loaded for the language `#1'. Using the pattern for}%
\typeout{** the default language instead.}%
\else
\language=\csname l@#1\endcsname
\fi
#2}}

\bibitem{fusion}
D.~J. Yeong \emph{et~al.}, ``{Sensor and sensor fusion technology in autonomous
  vehicles: A review},'' \emph{{Sensors}}, vol.~21, pp. 1--37, March 2021.

\bibitem{Angel}
M.~A. de~Miguel \emph{et~al.}, ``{High-Accuracy Patternless Calibration of
  Multiple 3-D LiDARs for Autonomous Vehicles},'' \emph{{IEEE Sensors
  Journal}}, vol.~23, 2023.

\bibitem{deepgmr}
W.~Yuan \emph{et~al.}, ``{DeepGMR: Learning Latent Gaussian Mixture Models for
  Registration},'' \emph{{ECCV}}, 2020.

\bibitem{gmmreg}
B.~Jian and B.~C. Vemuri, ``{Robust point set registration using Gaussian
  mixture models},'' \emph{{IEEE Transactions on Pattern Analysis and Machine
  Intelligence}}, vol.~33, pp. 1633--1645, 2011.

\bibitem{evangelidis}
G.~D. Evangelidis and R.~Horaud, ``{Joint Alignment of Multiple Point Sets with
  Batch and Incremental Expectation-Maximization},'' \emph{{IEEE Transactions
  on Pattern Analysis and Machine Intelligence}}, vol.~40, pp. 1397--1410, June
  2018.

\bibitem{hgmr}
B.~Eckart, K.~Kim, and J.~Kautz, ``{HGMR: Hierarchical Gaussian Mixtures for
  Adaptive 3D Registration},'' \emph{{ECCV}}, 2018.

\bibitem{besl}
P.~Besl and N.~McKay, ``{A Method for Registration of 3-D Shapes},''
  \emph{{IEEE Transactions on Pattern Analysis and Machine Intelligence}},
  vol.~14, pp. 239--256, 1992.

\bibitem{GICP}
A.~V. Segal, D.~Haehnel, and S.~Thrun, ``{Generalized-ICP},'' \emph{{Robotics:
  Science and Systems}}, vol.~2, 2009.

\bibitem{horaud}
\BIBentryALTinterwordspacing
R.~Horaud and F.~Dornaika, ``{Hand-eye Calibration},'' \emph{{The International
  Journal of Robotics Research}}, vol.~14, pp. 195--210, 1995. [Online].
  Available: \url{https://inria.hal.science/inria-00590039}
\BIBentrySTDinterwordspacing

\bibitem{lidarlink}
J.~Xu \emph{et~al.}, ``{LiDAR-Link: Observability-Aware Probabilistic
  Plane-Based Extrinsic Calibration for Non-Overlapping Solid-State LiDARs},''
  \emph{{IEEE Robotics and Automation Letters}}, vol.~9, pp. 2590--2597, March
  2024.

\bibitem{Zhang22}
J.~Zhang \emph{et~al.}, ``{LB-L2L-Calib: Accurate and Robust Extrinsic
  Calibration for Multiple 3D LiDARs with Long Baseline and Large Viewpoint
  Difference},'' \emph{{Proceedings - IEEE International Conference on Robotics
  and Automation}}, pp. 926--932, 2022.

\bibitem{Yoon21}
B.~H. Yoon, H.~W. Jeong, and K.~S. Choi, ``{Targetless Multiple Camera-LiDAR
  Extrinsic Calibration using Object Pose Estimation},'' \emph{{Proceedings -
  IEEE International Conference on Robotics and Automation}}, vol. 2021-May,
  pp. 13\,377--13\,383, 2021.

\bibitem{Li22}
Z.~Li \emph{et~al.}, ``{Extrinsic Calibration of a 2D Laser Rangefinder and a
  Depth-camera Using an Orthogonal Trihedron},'' \emph{{IEEE International
  Conference on Intelligent Robots and Systems}}, vol. 2022-October, pp.
  6264--6269, 2022.

\bibitem{Kim21}
J.~Kim \emph{et~al.}, ``{Automated Extrinsic Calibration for 3D LiDARs with
  Range Offset Correction using an Arbitrary Planar Board},''
  \emph{{Proceedings - IEEE International Conference on Robotics and
  Automation}}, vol. 2021-May, pp. 5082--5088, 2021.

\bibitem{Jiao}
J.~Jiao \emph{et~al.}, ``{A Novel Dual-Lidar Calibration Algorithm Using Planar
  Surfaces},'' \emph{{2019 IEEE Intelligent Vehicles Symposium (IV)}}, 2019.

\bibitem{Hu19}
H.~Hu and othersr, ``{TEScalib: Targetless Extrinsic Self-Calibration of LiDAR
  and Stereo Camera for Automated Driving Vehicles with Uncertainty
  Analysis},'' \emph{{IEEE International Conference on Intelligent Robots and
  Systems}}, vol. 2022-October, pp. 6256--6263, 2022.

\bibitem{Wei22}
P.~Wei \emph{et~al.}, ``{CROON: Automatic Multi-LiDAR Calibration and
  Refinement Method in Road Scene},'' \emph{{IEEE International Conference on
  Intelligent Robots and Systems}}, vol. 2022-October, pp. 12\,857--12\,863,
  2022.

\bibitem{Serafin15}
J.~Serafin and G.~Grisetti, ``{NICP: Dense Normal Based Point Cloud
  Registration},'' \emph{{2015 IEEE/RSJ International Conference on Intelligent
  Robots and Systems (IROS)}}, 2015.

\bibitem{Heide18}
N.~Heide, T.~Emter, and J.~Petereit, ``{Calibration of multiple 3D LiDAR
  sensors to a common vehicle frame},'' \emph{{ISR 2018; 50th International
  Symposium on Robotics}}, 2018.

\bibitem{emicp}
S.~Granger and X.~Pennec, ``{Multi-scale EM-ICP: A Fast and Robust Approach for
  Surface Registration},'' \emph{{ECCV}}, pp. 69--73, 2002.

\bibitem{emalgo}
A.~Dempster, N.~Laird, and D.~Rubin, ``{Maximum likelihood from incomplete data
  via the em algorithm},'' \emph{{Journal of the Royal Statistical Society}},
  pp. 1--38, 1977.

\bibitem{ndt}
T.~D. Stoyanov \emph{et~al.}, ``{Fast and accurate scan registration through
  minimization of the distance between compact 3D NDT representations},''
  \emph{{International Journal of Robotics Research}}, 2012.

\bibitem{mlmd}
B.~Eckart \emph{et~al.}, ``{MLMD: Maximum Likelihood Mixture Decoupling for
  Fast and Accurate Point Cloud Registration},'' \emph{{Proceedings - 2015
  International Conference on 3D Vision, 3DV 2015}}, pp. 241--249, Nov 2015.

\bibitem{craig}
J.~J. Craig, ``{Introduction to Robotics Mechanics and Control Third
  Edition},'' \emph{{Pearson Prentice Hall}}, 2005.

\bibitem{edgar}
\BIBentryALTinterwordspacing
P.~Karle \emph{et~al.}, ``{EDGAR: An Autonomous Driving Research Platform –
  From Feature Development to Real-World Application},'' Jan 2024. [Online].
  Available: \url{https://arxiv.org/abs/2309.15492}
\BIBentrySTDinterwordspacing

\bibitem{carla}
\BIBentryALTinterwordspacing
A.~Dosovitskiy \emph{et~al.}, ``{CARLA: An Open Urban Driving Simulator},'' Jan
  2017. [Online]. Available: \url{https://arxiv.org/abs/1711.03938}
\BIBentrySTDinterwordspacing

\end{thebibliography}
\bibliographystyle{IEEEtran}

\end{document}